\documentclass[twocolumn]{article}
\usepackage[latin1]{inputenc}
\usepackage{amsmath}
\usepackage{amsfonts}
\usepackage{amssymb}
\usepackage{mathrsfs}
\usepackage{multirow}
\usepackage{balance}
\usepackage{tcolorbox}

\usepackage{blindtext}

\usepackage{MnSymbol}

\usepackage[cal=boondox,scr=boondoxo]{mathalfa}

\usepackage{float}
\usepackage{fancybox,graphicx}
\usepackage{subfig}
\usepackage{caption}
\usepackage{color}
\usepackage{authblk}

\usepackage[colorlinks]{hyperref}

\usepackage{hyperref}

\newcommand{\doi}[1]{{doi:~\href{https://doi.org/#1}{\nolinkurl{#1}}}\rmFullStop}

\newcommand*{\rmFullStop}{\rmifnextchar{.}{}{}}

\makeatletter
\newcommand{\rmifnextchar}[3]{%
  \begingroup
  \ltx@LocToksA{\endgroup#2}%
  \ltx@LocToksB{\endgroup#3}%
  \ltx@ifnextchar{#1}{%
    \def\next{\the\ltx@LocToksA}%
    \afterassignment\next
    \let\scratch= %
  }{%
    \the\ltx@LocToksB
  }%
}
\makeatother 

\usepackage{accents}
\usepackage[titletoc,title]{appendix}
\usepackage{cite}

\usepackage{mathtools}

\usepackage[top=2in, bottom=1.5in, left=1in, right=1in]{geometry}

\usepackage{stackengine}

\title{AraSTEM: A Native Arabic Multiple Choice Question Benchmark for Evaluating LLMs Knowledge In STEM Subjects} 
 
\author{Ahmad Mustapha}
\author{Hadi Al-Khansa}
\author{Hadi Al-Mubasher}
\author{Aya Mourad}
\author{Ranam Hamoud}
\author{Hasan El-Husseini}
\author{Marwah Al-Sakkaf}
\author{Mariette Awad}

\affil{American University of Beirut, Lebanon}

\begin{document}

\maketitle

\begin{abstract}

Large Language Models (LLMs) have shown remarkable capabilities, not only in generating human-like text, but also in acquiring knowledge. This highlights the need to go beyond the typical Natural Language Processing downstream benchmarks and asses the various aspects of LLMs including knowledge and reasoning. Numerous benchmarks have been developed to evaluate LLMs knowledge, but they predominantly focus on the English language. Given that many LLMs are multilingual, relying solely on benchmarking  English knowledge is insufficient. To address this issue, we introduce AraSTEM, a new Arabic multiple-choice question dataset aimed at evaluating LLMs knowledge in STEM subjects. The dataset spans a range of topics at different levels which requires models to demonstrate a deep understanding of scientific Arabic in order to achieve high accuracy. Our findings show that publicly available models of varying sizes struggle with this dataset, and underscores the need for more localized language models. The dataset is freely accessible on Hugging Face.

Keywords: LLMs, Arabic Language, STEM, Reasoning.
\end{abstract}

\section{Introduction}

Language models are traditionally evaluated on various Natural Language Processing (NLP) tasks such as text generation, sentiment analysis, translation, and Part of Speech (POS) tagging, using metrics like the BLEU score. However, with the advent of Large Language Models (LLMs), these metrics and benchmarks have become outdated. Unlike conventional language models, LLMs exhibit zero-shot and few-shot learning abilities, enabling them to become better across many existing benchmarks. These models have demonstrated the ability to perform tasks that require not only language comprehension but also knowledge acquisition and reasoning. In essence, LLMs function as multi-faceted agents, and their capabilities must be assessed from multiple angles, including knowledge, reasoning, and alignment with human preferences.

\begin{figure*}[ht]
\centering
\includegraphics[width=0.9\textwidth]{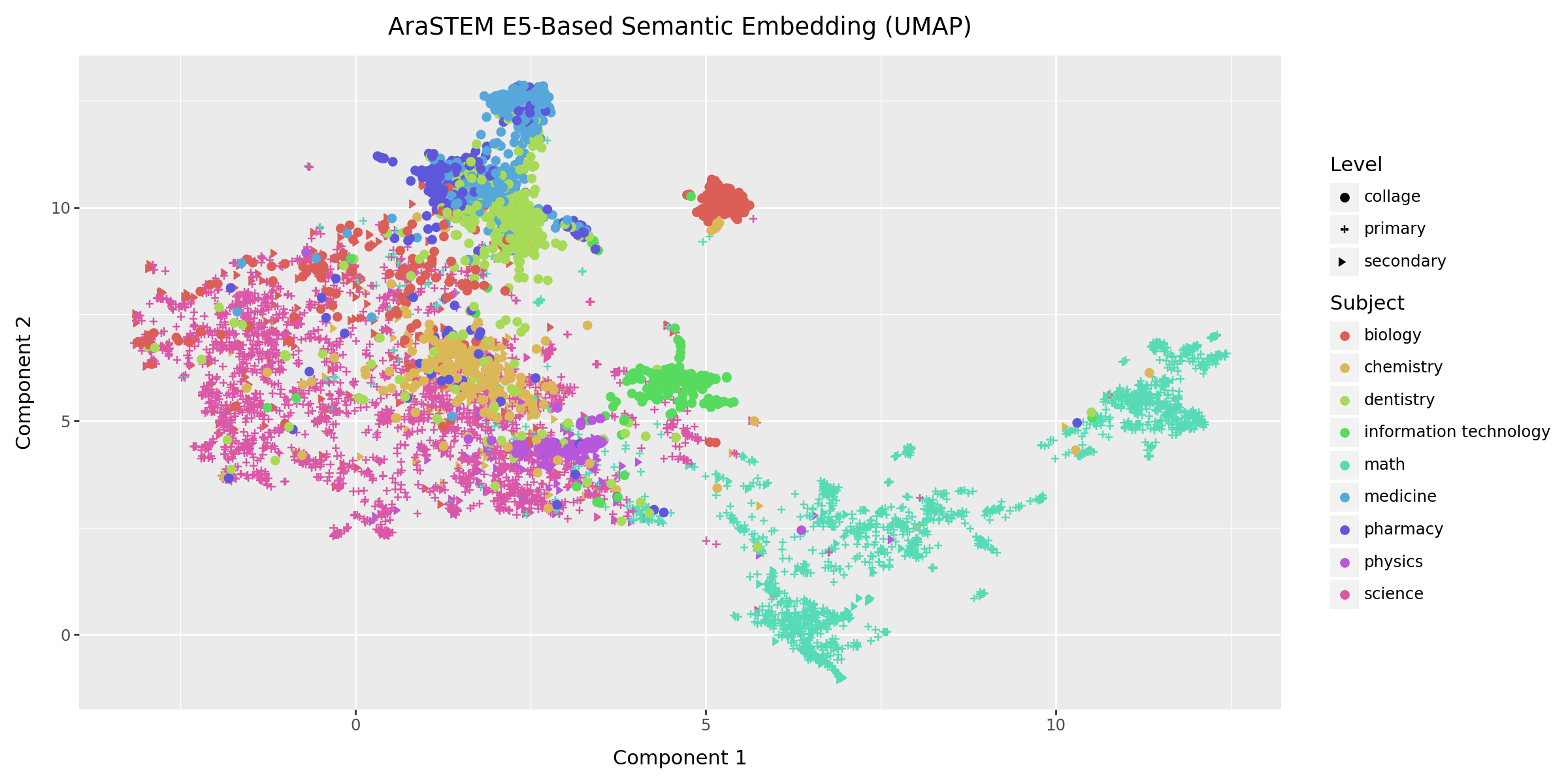}
    \captionsetup{justification=centering,margin=1.5cm}
    \caption{Semantic Embedding of AraSTEM based on E5 multilingual embedding model. Projected using UMAP}
\label{fig:arastem-semantic-embedding}
\end{figure*}

While numerous benchmarks have been proposed, the majority of them is English-centric, despite the fact that several open-source and proprietary models claim robust multilingual capabilities. To make these models more accessible for various languages, it is essential to create language-specific benchmarks that evaluate their performance in these languages. One language that lacks sufficient and comprehensive evaluation benchmarks is Arabic. Arabic ranks fifth in the world's league table of languages, with over 200 million native Arabic speakers worldwide. It is supported in both open-source and closed-source models. For instance, both Llama 3 and ChatGPT 4 support it. The BLOOMZ model \cite{Muennighoff2022} and the Aya model \cite{Singh2024} reported that 5-6\% of the training data was in Arabic. Additionally, Jais \cite{Sengupta2023}, the first Arabic-specific LLM, is reported to have been trained on a dataset where approximately 28\% of the data is in Arabic. 

To resolve the desertification of Arabic language benchmarks, we introduce AraSTEM, an Arabic dataset designed to serve as a benchmark for evaluating Arabic knowledge in Science, Technology, Engineering, and Math (STEM) fields based on native content. The dataset consists of a total of 11637 multiple-choice questions (MCQs) across subjects such as Math, Science, Physics, Biology, Chemistry, Computer Science, and Medicine. The questions range from elementary school level to professional or college-level difficulty.  We also assessed the performance of various open-source LLMs of different sizes on this dataset in a zero-shot setting. Our findings reveal that there still a room for improvement in many of the SOTA language models. The dataset is available on Hugging Face \footnote{Available post publication}.

The rest of this manuscript is organized as following: 
Section \ref{related-work} discusses related work. Section \ref{dataset} presents the proposed dataset, the collection process, and it's details. Section \ref{experiments} explains the experiment's setup and results while section \ref{conclusions} concludes the work with follow on recommendations.

\section{Related Work} \label{related-work}

Because LLMs have shown superior performance on many conventional NLP benchmarks, many proposed contemporary benchmarks that are more challenging for LLMs. Hendrycks \cite{Hendrycks2020} developed the Massive Multitask Language Understanding (MMLU) benchmark which tests against world knowledge along 57 different topics. Zellers \cite{Zellers2019} on the other hand produced a benchmark that tests human common sense by generating  entries using adversarial filtering which makes them challenging for models but easy for humans. Sakaguchi \cite{Sakaguchi2019} compiled WinoGrande which is a scaled version of the Winograd benchmark where the model is asked to resolve pronouns in challenging text entries that can't be resolved without language understanding. Zheng \cite{Zheng2023} proposed the MT-Bench benchmark which is composed of open-ended multi-turn questions that test conversational aspects in LLMs. Other benchmarks include MATH \cite{Hendrycks2021}, BIG-Bench \cite{Suzgun2022}, DROP \cite{Dua2019}, MMLU-Pro \cite{Wang2024}, GPQA \cite{Rein2023}, and IFEval \cite{Zhou2023}.

Regarding the Arabic language only a handful of benchmarks exists. Openai recently translated, using human translators, the MMLU dataset \cite{Hendrycks2020} into 14 different languages including Arabic under the name Multilingual Massive Multitask Language Understanding (MMMLU) \cite{MMMLU}. Contrary to our data MMMLU is not native. On the other hand, Koto proposed ArabicMMLU \cite{Koto2024} a total of 14575 MCQ questions. Contrary to our data the STEM part of the ArabicMMLU corresponded to 20\% of the data while ours is STEM focused.

On a different note, several studies have explored the Arabic language from various angles. Alharbi \cite{alharbi} demonstrated that deep neural networks, optimized through genetic algorithms, significantly boost the performance of Arabic sentiment classification. Hajj \cite{Hajj} presented a cortical algorithm-based model for Arabic Automatic Speech Recognition (ASR). Alsayadi \cite{alsayadi} integrated CNNs with Long Short-Term Memory (LSTM) networks to improve recognition of non-diacritized Arabic speech. Bahatia \cite{Bhatia}  proposed a multimodal language model for both Arabic optical and handwriting recognition.

Since our dataset includes medical questions, some research has focused on similar medical question-related tasks. El Zini \cite{elziniOSCE} introduced a corpus and a deep learning framework for Objective Structured Clinical Examinations (OSCE). Qiu et al. \cite{Qiu2019} developed a deep learning model to predict the difficulty level of medical exams.

\section{The Dataset} \label{dataset}

The AraSTEM dataset contains 11,637 multiple-choice questions covering a wide range of knowledge areas. It includes questions from elementary and secondary level math and science, as well as advanced topics in biology, physics, and medicine. The dataset was compiled from various sources, and in response to the issues recently raised about data traceability and governance \cite{Longpre}, we opted to cite the source for each individual question.

\subsection{Data Sources and Data Collection}

The data was collected from multiple sources, each employing a distinct collection process. Table \ref{table:sources-summary} outlines the sources, the number of questions extracted from each, and a description of the extraction process used.

\begin{table*}[ht]
\centering
\begin{tabular}{|l|l|l|r|}
\hline
\textbf{Source} & \textbf{Type} & \textbf{Extraction Method} & \textbf{Count} \\ \hline
beadaya.com & MCQ Website & Scraping & 7721 \\ \hline
www.mehe.gov.lb & PDF File & LLM Extraction & 1806 \\ \hline
platform.almanhal.com & Book & Manual Extraction & 914 \\ \hline
eqiyas.com & MCQ Website & Manual Extraction & 414 \\ \hline
www.alloschool.com & MCQ Website & Scraping & 277 \\ \hline
manara.edu.sy & PDF File & Manual Extraction & 137 \\ \hline
slideshare.net & PDF File & Manual Extraction & 125 \\ \hline
faculty.ksu.edu.sa & PDF File & Manual Extraction & 96 \\ \hline
kau.edu.sa & MCQ Website & Manual Extraction & 79 \\ \hline
awa2el.net & MCQ Website & Manual Extraction & 69 \\ \hline
\end{tabular}
\captionsetup{justification=centering,margin=2cm}
\caption{A summary of AraSTEM data collection figures}
\label{table:sources-summary}
\end{table*}

\begin{figure}[h]
\centering
\captionsetup{justification=centering,margin=0.3cm}
\includegraphics[width=0.45\textwidth]{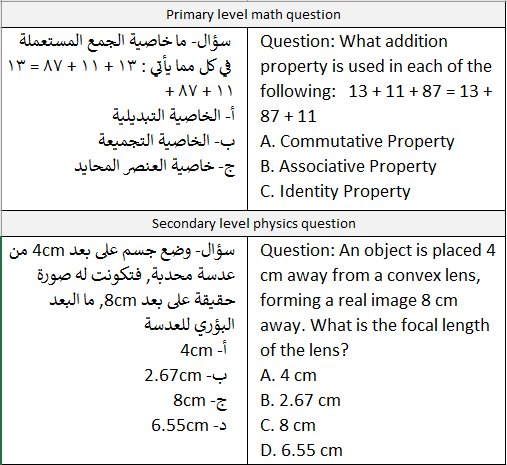}
    \caption{A sample from AraSTEM questions corresponding to primary and secondary levels}
\label{fig:primary-secondary-mcq-example}
\end{figure}

\textbf{Scraping.} We scraped two publicly accessible MCQ websites, beadaya.com and alloschool.com, using Python scripts with libraries like BeautifulSoup and Requests. Many of the questions collected from these sites included images and mathematical equations, which were removed during the cleaning process. To maintain traceability, each question from these sources includes a link in the dataset that directs to its original source. The scraped questions cover a range of topics, from simple science questions for primary school students to more advanced physics and chemistry questions at the secondary school level. Figure \ref{fig:primary-secondary-mcq-example} provides examples of questions from both primary and secondary levels.

\begin{figure}[h]
\centering
\captionsetup{justification=centering,margin=0.3cm}
\includegraphics[width=0.5\textwidth]{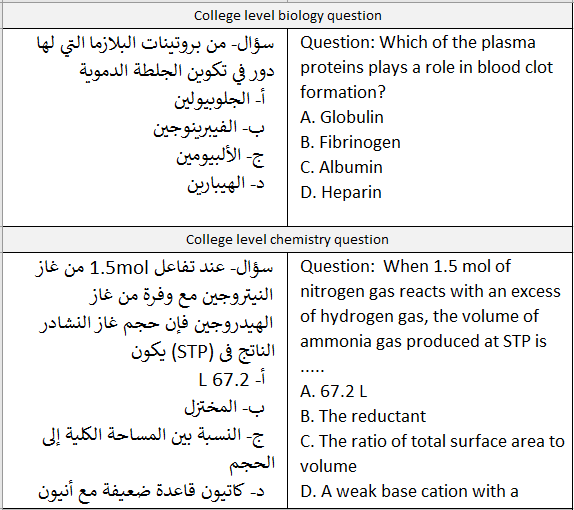}
    \caption{A sample from AraSTEM questions featuring college level ones in both chemistry and biology}
\label{fig:college-mcq-example}
\end{figure}

\textbf{Manual Extraction.} A portion of the questions was manually collected from two reference books that contain thousands of questions and answers in biology and chemistry. While the majority of these questions were not originally in multiple-choice format, we were able to extract some MCQs. Moreover, we generated additional MCQs by transforming other types of questions. For example, "definition" questions were turned into MCQs by rearranging possible answers, and "fill in the blanks" questions were similarly converted by offering multiple answer options, including the correct one. To ensure traceability, each entry in the dataset includes a link to a public version of the book. Additionally, several publicly available MCQs were extracted manually by the authors from various online sources, primarily featuring college-level questions in subjects like math, physics, chemistry, and information technology. Figure \ref{fig:college-mcq-example} provides an example of these college-level questions.

\begin{figure}[h]
\centering
\captionsetup{justification=centering,margin=0.3cm}
\includegraphics[width=0.45\textwidth]{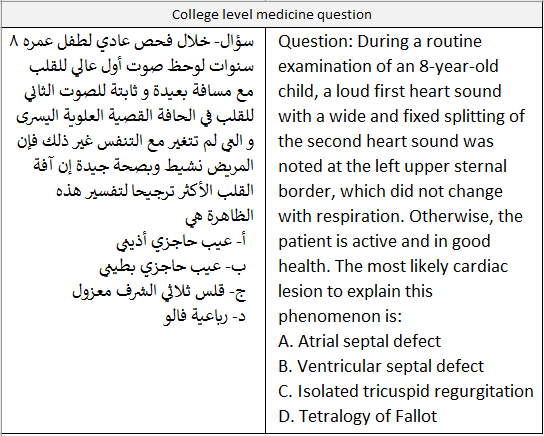}
    \caption{A sample from AraSTEM questions featuring college-level medicine question}
\label{fig:medicine-mcq-example}
\end{figure}

\textbf{LLM Extraction.} A portion of the questions was extracted from PDF files using ChatGPT 4. We conducted a small validation study to assess its performance as an OCR engine and found that, while not optimal, it was sufficiently effective for extracting large amounts of text if combined with manual efforts. Hundreds of pages from medical colloquiums were processed by feeding them as images to ChatGPT 4 for text extraction. After automatic extraction, the authors manually proofread and corrected the data. This extracted content consisted of medical questions in fields such as medicine, pharmacy, and dentistry. The inclusion of these questions made the dataset both more challenging and comprehensive. Figure \ref{fig:medicine-mcq-example} presents an example of college-level medical questions.

\begin{figure*}[ht]
\centering
\captionsetup{justification=centering,margin=0.3cm}
\includegraphics[width=0.9\textwidth]{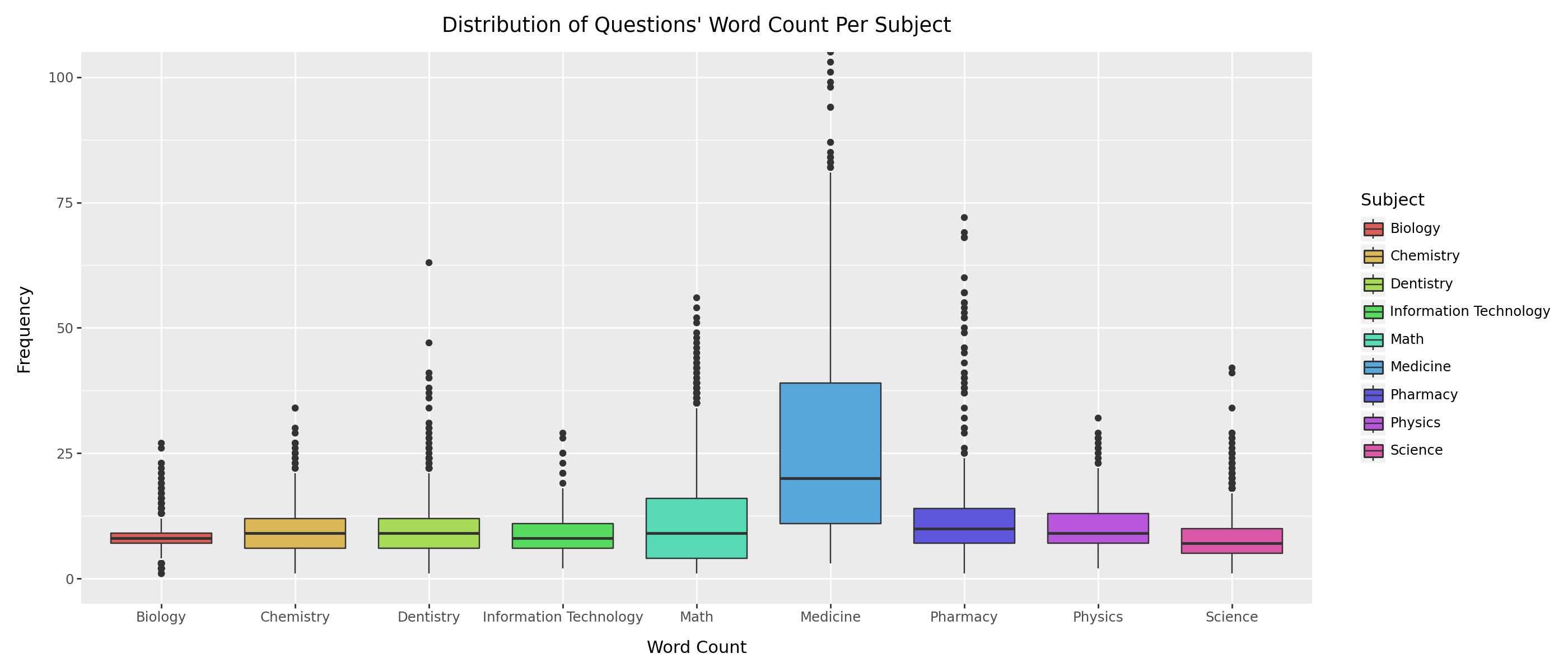}
    \caption{The distribution of AraSTEM question's word count presented per subject}
\label{fig:number-of-words-per-question}
\end{figure*}

\subsection{Data Characteristics}

\textbf{Distribution Per Level and Subject.} The questions in the dataset are classified into three educational levels: primary, secondary, and college. And it encompasses a broad array of subjects, including math, science, physics, chemistry, biology, information technology, dentistry, pharmacy, and medicine. The "science" category pertains specifically to primary-level questions that includes physics, chemistry, and biology but was labeled as "science" from the source. Table \ref{table:distribution-of-questions} illustrates the distribution of questions by level and subject. While a substantial portion of the dataset comprises primary-level questions, this does not diminish its inclusivity or the challenging nature of its content.

\textbf{Distribution Per Number of Options.} The number of options per question varies. Some questions provide four options, while others offer only two. Table \ref{table:number-of-options} displays the distribution of options across the questions.

\textbf{Word Count Distribution.} An important aspect to note is the length of the questions. To evaluate this, we calculated the word count for each question and plotted the distribution of these counts. Figure \ref{fig:number-of-words-per-question} illustrates the results. While some questions can contain up to 170 words, most fall within the range of 10 to 15 words.The medicine subject has the longest questions, followed closely by math.

\begin{table}[h!]
\centering
\captionsetup{justification=centering,margin=0.3cm}
\begin{tabular}{|c|c|c|}
\hline
\textbf{Level}   & \textbf{Subject}              & \textbf{Count} \\ \hline
\multirow{2}{*}{Primary}   & Math                 & 3220  \\ \cline{2-3} 
                          & Science              & 4250  \\ \hline
\multirow{5}{*}{Secondary} & Biology              & 322   \\ \cline{2-3} 
                          & Chemistry            & 240   \\ \cline{2-3} 
                          & Math                 & 84    \\ \cline{2-3} 
                          & Physics              & 181   \\ \cline{2-3} 
                          & Science              & 115   \\ \hline
\multirow{7}{*}{College}   & Biology              & 588   \\ \cline{2-3} 
                          & Chemistry            & 326   \\ \cline{2-3} 
                          & Dentistry            & 657   \\ \cline{2-3} 
                          & Information Technology & 369   \\ \cline{2-3} 
                          & Medicine             & 721   \\ \cline{2-3} 
                          & Pharmacy             & 428   \\ \cline{2-3} 
                          & Physics              & 137   \\ \hline
\end{tabular}
\caption{Count of AraSTEM questions grouped by levels and distributed over subjects}
\label{table:distribution-of-questions}
\end{table} 

\begin{figure}[H]
\centering
\captionsetup{justification=centering,margin=0.3cm}
\includegraphics[width=0.5\textwidth]{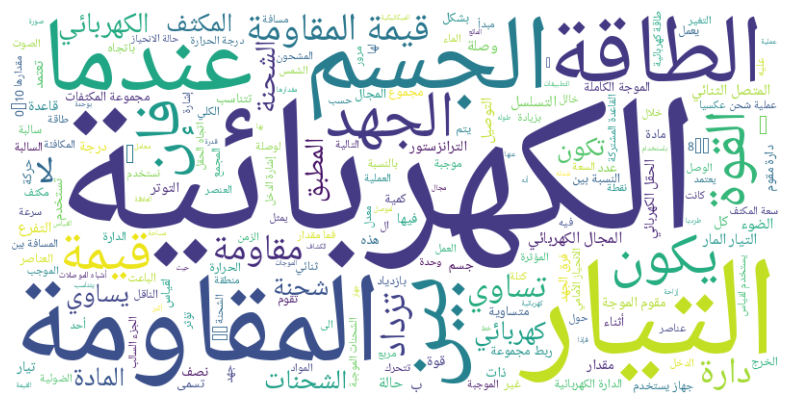}
 \caption{The word cloud illustration of AraSTEM questions of the physics subject}
\label{fig:dentistry-word-count}
\end{figure}

\vspace{1cm}

\textbf{Semantic Distribution.} To examine the dataset's diversity from a semantic perspective, we utilized the multilingual text embedding model E5-Large \cite{wang2024multilingual} to project all the questions into an embedding. Figure \ref{fig:arastem-semantic-embedding} displays the results, where the embedding is visualized in two dimensions using UMAP \cite{umap}. The distribution indicates a clear separation between mathematical questions and the other categories, which is anticipated due to the inclusion of mathematical terms and equations. Interestingly, the math group is further divided into three distinct clusters, showcasing the dataset's diversity. Additionally, questions from dentistry, pharmacy, and medicine form a nearby cluster, while the science questions represent a large, semantically varied group. Other subjects also tend to cluster separately, emphasizing the semantic distinctions across the various topics. 

To further illustrate the dataset's diversity, we computed and plotted word clouds for questions from different subjects. The word clouds clearly reveal the distinct nature of the questions within the dataset and highlight the primary focus and the nature of questions of each subject. Figure \ref{fig:dentistry-word-count} presents the cloud for the physics subject. A larger list of figures is presented in the appendix section \ref{app:word clouds}.

\textbf{Unknown Vocabulary.} LLMs are trained with a vocabulary based on the data they learn from. This raises a question: Does our dataset have words that aren't in the model's vocabulary? To find out, we tested the entire dataset with different models' tokenizers and counted how many unknown words each model detected. Some models fully covered the dataset's vocabulary, including Jais 13B chat (which supports both Arabic and English), Llama 2 7B chat, AceGPT 7B chat, and BLOOM 560M. Other models found only a few unknown words. For instance, XGLM 1.7B found 325 unknown words, while AraT5 base found 698. Since our dataset contains some rare words, these results show that many models have strong coverage of Arabic, even if they don't officially support the language.

\begin{table}[htbp]
    \centering
    \captionsetup{justification=centering,margin=0.3cm}
    \begin{tabular}{|c|c|}
        \hline
        \textbf{Number of Options} & \textbf{Count} \\ \hline
        2 & 2562 \\ \hline
        3 & 3808 \\ \hline
        4 & 5268 \\ \hline
    \end{tabular}
    \caption{The number of questions' options and their counts in the AraSTEM dataset}
    \label{table:number-of-options}
\end{table}

\section{Experiments} \label{experiments}

To evaluate the difficulty of the dataset, we tested the performance of several language models. These models ranged from smaller ones with around 500 million parameters to medium-sized models with 7 billion parameters, and up to larger models with 30 to 40 billion parameters.

\tcbset{
    colback=black!5!white, 
    colframe=black!75!black,
    fonttitle=\bfseries,
    fontupper=\small, 
    title=Prompt,
    boxrule=0.5mm,
    width=0.5\textwidth,
    arc=2mm,
    auto outer arc,
    boxsep=1mm,
}

\begin{figure}[htbp]
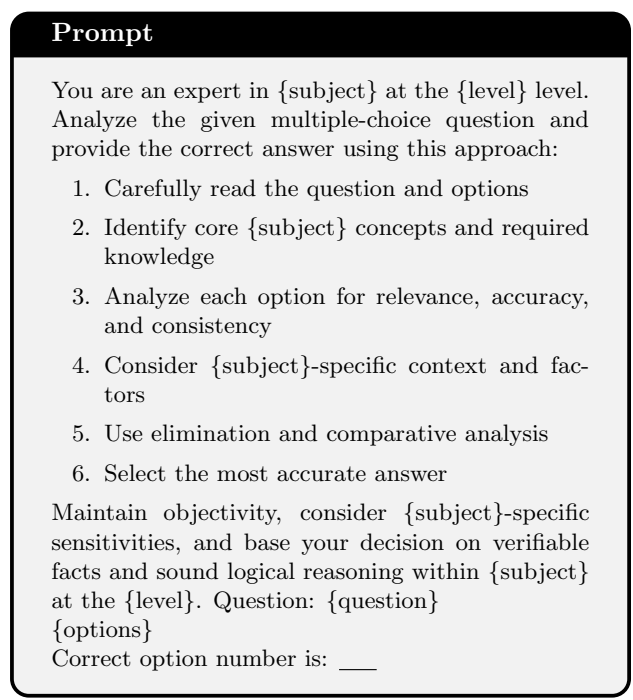

    \centering
    \captionsetup{justification=centering,margin=0.3cm}
    \begin{tcolorbox}
        You are an expert in \{subject\} at the \{level\} level. Analyze the given multiple-choice question and provide the correct answer using this approach:
        \begin{enumerate}
            \item Carefully read the question and options
            \item Identify core \{subject\} concepts and required knowledge
            \item Analyze each option for relevance, accuracy, and consistency
            \item Consider \{subject\}-specific context and factors
            \item Use elimination and comparative analysis
            \item Select the most accurate answer
        \end{enumerate}
        Maintain objectivity, consider \{subject\}-specific sensitivities, and base your decision on verifiable facts and sound logical reasoning within \{subject\} at the \{level\}.
        Question:
        \{question\} \\
        \{options\} \\
        Correct option number is: \underline{\hspace{0.5cm}}
    \end{tcolorbox}
\caption{The prompt used in the evaluation of several models performance on AraSTEM dataset}
\label{fig:cot-prompt}
\end{figure}

Each model was tasked with answering the dataset's questions by selecting from the available options. The models' choices were extracted as the probability of predicting each one of the following tokens "A", "B", "C", or "D" representing the possible answer choices. Alongside the predictions the models' confidence have been also recorded. The confidence were calculated by applying a softmax function over the models' output logits of the target tokens. If a question and its prompt exceeded the models' maximum input size, the prompt was truncated accordingly.

\subsection{Prompt Engineering}
We designed the prompt based on recent research findings. First, we wrote the prompt in English, following Koto \cite{Koto2024}, who found that using English for the body of the prompt - even for Arabic questions - improves performance across multiple models. Second, we applied Chain-of-Thought (CoT) prompting, as suggested by the latest research \cite{cot}. In this approach, the model is instructed to carefully analyze the question, identify the key concepts related to the subject, and then proceed to answer. The exact prompt format is shown in Figure \ref{fig:cot-prompt}.

\subsection{Zero Shot Performance} \label{sec:zero-shot}

Table \ref{table:zero_shot_performance} presents the results of this experimental setup. The best-performing model, on average, is Jais 30B Chat, achieving an accuracy of 56\%, followed closely by the Jais 13B chat and Llama 3.1 8B instruct models. This high score highlights the challenging nature of our dataset.

In the table, we bolded the top four accuracies per subject for each model group. A clear trend emerges: Jais, Llama 3.1, Bloomz, and AceGPT consistently achieve the highest accuracies across subjects, positioning them as the top-performing group on average. This pattern is notable, as Bloomz and AceGPT models outperform popular models like Llama 2 and Falcon. The results suggest the value of incorporating substantial Arabic text in the models' training datasets; indeed, the top performers have all cited Arabic as part of their training data except for Lama 3.1. Specifically, Jais reported that 28\% of its training data is in Arabic, Bloomz 6\%, and AceGPT \cite{huang-etal-2024-acegpt} 64\%. Llama 3.1 \cite{grattafiori2024llama3herdmodels} didn't disclose the size of the Arabic corpus in the training set.

\begin{table}[]
    \centering
    \captionsetup{justification=centering,margin=0.3cm}
    \renewcommand{\arraystretch}{1.3}
    \begin{tabular}{l|cccc}
    \hline
    \textbf{Model} & \rotatebox{90}{\textbf{Primary}} & \rotatebox{90}{\textbf{Secondary}} & \rotatebox{90}{\textbf{Collage}} & \rotatebox{90}{\textbf{Average}} \\
    \hline
    Random Guess          & 0.37 & 0.27 & 0.27 & 0.30 \\
    Arat5V2 Base          & 0.38 & 0.30 & 0.24 & 0.31 \\
    Aragpt2 Base          & 0.40 & 0.32 & 0.24 & 0.32 \\
    Mt0 Large             & 0.40 & 0.32 & 0.23 & 0.32 \\
    Xglm 7.5B             & 0.34 & 0.26 & 0.25 & 0.28 \\
    Bloomz 7B1            & 0.49 & 0.40 & 0.42 & 0.44 \\
    Acegpt 13B Chat       & 0.56 & 0.43 & 0.40 & 0.46 \\
    Llama 3.1 8B Instruct & 0.61 & 0.55 & 0.46 & 0.54 \\
    Jais 30B Chat V3      & 0.62 & 0.58 & 0.53 & 0.58 \\
    Falcon 40B Instruct   & 0.41 & 0.31 & 0.29 & 0.34 \\
    \hline
    \end{tabular}
    \caption{Performance of selected models over the AraSTEM dataset. The results are reported per educational level.}
    \label{table:zsp-level}
\end{table}

Among the top performing models, the easiest subjects were Science, Information Technology (IT), Chemistry, and Biology. Conversely, the most challenging subjects were Dentistry, Medicine, Pharmacy, and Math.

\begin{table*}[]
    \centering
    \captionsetup{justification=centering,margin=2.8cm}
    \renewcommand{\arraystretch}{1.3}
    \small
    \begin{tabular}{|l|c|c|c|c|c|c|c|c|c|c|}
    \hline
    \textbf{Model} & 
    \rotatebox{90}{\textbf{Biology}} & 
    \rotatebox{90}{\textbf{Chemistry}} & 
    \rotatebox{90}{\textbf{Dentistry}} & 
    \rotatebox{90}{\textbf{IT}} & 
    \rotatebox{90}{\textbf{Math}} & 
    \rotatebox{90}{\textbf{Medicine}} & 
    \rotatebox{90}{\textbf{Pharmacy}} & 
    \rotatebox{90}{\textbf{Physics}} & 
    \rotatebox{90}{\textbf{Science}} & 
    \rotatebox{90}{\textbf{Average}} \\
    \hline
        Random Guess & 0.25 & 0.31 & 0.25 & 0.29 & 0.34 & 0.25 & 0.25 & 0.32 & 0.39 & 0.29 \\
        \hline
        Arat5 Base & 0.22 & 0.16 & 0.22 & 0.18 & 0.03 & 0.19 & 0.24 & 0.15 & 0.06 & 0.16 \\
        Arat5V2 Base & 0.25 & 0.26 & 0.21 & 0.27 & 0.34 & 0.22 & 0.25 & 0.27 & 0.41 & 0.28 \\
        Aragpt2 Base & 0.25 & 0.29 & 0.22 & 0.26 & 0.35 & 0.20 & 0.23 & 0.31 & 0.44 & 0.28 \\
        \hline
        Mt0 Small & 0.26 & 0.29 & 0.21 & 0.28 & 0.35 & 0.20 & 0.23 & 0.30 & 0.42 & 0.28 \\
        Mt0 Base & 0.27 & 0.29 & 0.21 & 0.28 & 0.37 & 0.20 & 0.23 & 0.34 & 0.43 & 0.29 \\
        Mt0 Large & 0.25 & 0.30 & 0.22 & 0.26 & 0.35 & 0.20 & 0.19 & 0.31 & 0.44 & 0.28 \\
        \hline
        Xglm 1.7B & 0.25 & 0.30 & 0.22 & 0.27 & 0.35 & 0.20 & 0.23 & 0.30 & 0.44 & 0.28 \\
        Xglm 2.9B & 0.25 & 0.29 & 0.22 & 0.26 & 0.35 & 0.20 & 0.23 & 0.31 & 0.44 & 0.28 \\
        Xglm 4.5B & 0.26 & 0.28 & 0.21 & 0.30 & 0.35 & 0.20 & 0.24 & 0.30 & 0.44 & 0.29 \\
        Xglm 7.5B & 0.24 & 0.30 & 0.23 & 0.29 & 0.33 & 0.20 & 0.23 & 0.33 & 0.34 & 0.28 \\
        \hline
        Bloomz 560M & 0.22 & 0.33 & 0.19 & 0.30 & 0.31 & 0.26 & 0.25 & 0.30 & 0.34 & 0.28 \\
        Bloomz 1B1 & 0.24 & 0.30 & 0.19 & 0.25 & 0.29 & 0.27 & 0.28 & 0.26 & 0.22 & 0.26 \\
        Bloomz 1B7 & 0.41 & 0.44 & 0.29 & 0.46 & 0.34 & 0.26 & 0.26 & 0.40 & 0.52 & 0.38 \\
        Bloomz 3B & 0.41 & 0.47 & 0.29 & 0.55 & 0.36 & 0.28 & 0.28 & 0.44 & 0.58 & 0.41 \\
        Bloomz 7B1 & \textbf{0.50} & \textbf{0.52} & \textbf{0.31} & \textbf{0.55} & \textbf{0.36} & \textbf{0.29} & \textbf{0.34} & \textbf{0.47} & \textbf{0.59} & \textbf{0.44} \\
        \hline
        Acegpt 7B & 0.36 & 0.38 & 0.23 & 0.41 & 0.33 & 0.27 & 0.26 & 0.33 & 0.48 & 0.34 \\
        Acegpt 7B Chat & 0.45 & 0.38 & 0.25 & 0.53 & 0.36 & 0.23 & 0.26 & 0.38 & 0.60 & 0.38 \\
        Acegpt 13B & 0.49 & 0.44 & 0.28 & 0.55 & 0.36 & 0.28 & 0.29 & 0.36 & 0.62 & 0.41 \\
        Acegpt 13B Chat & \textbf{0.53} & \textbf{0.45} & \textbf{0.28} & \textbf{0.60} & \textbf{0.38} & \textbf{0.28} & \textbf{0.32} & \textbf{0.42} & \textbf{0.69} & \textbf{0.44} \\
        \hline
        Llama 2 7B & 0.26 & 0.34 & 0.25 & 0.34 & 0.32 & 0.21 & 0.27 & 0.31 & 0.39 & 0.30 \\
        Llama 2 7B Chat & 0.33 & 0.38 & 0.24 & 0.36 & 0.30 & 0.22 & 0.27 & 0.40 & 0.36 & 0.32 \\
        Llama 2 13B & 0.28 & 0.31 & 0.22 & 0.40 & 0.34 & 0.21 & 0.24 & 0.32 & 0.46 & 0.31 \\
        Llama 2 13B Chat & 0.33 & 0.32 & 0.23 & 0.39 & 0.33 & 0.26 & 0.27 & 0.30 & 0.43 & 0.32 \\
        Llama 3.1 8B & 0.57 & 0.47 & 0.32 & 0.59 & 0.41 & 0.29 & 0.40 & 0.46 & 0.67 & 0.46 \\
        Llama 3.1 8B Instruct & \textbf{0.59} & \textbf{0.53} & \textbf{0.34} & \textbf{0.66} & \textbf{0.46} & \textbf{0.32} & \textbf{0.43} & \textbf{0.50} & \textbf{0.72} & \textbf{0.51} \\
        \hline
        Jais 13B & 0.38 & 0.37 & 0.25 & 0.44 & 0.35 & 0.26 & 0.27 & 0.38 & 0.43 & 0.35 \\
        Jais 13B Chat & 0.63 & 0.61 & 0.33 & 0.66 & 0.37 & 0.34 & 0.42 & 0.51 & 0.73 & 0.51 \\
        Jais 30B V3 & 0.55 & 0.50 & 0.29 & 0.57 & 0.37 & 0.25 & 0.34 & 0.43 & 0.66 & 0.44 \\
        Jais 30B Chat V3 & \textbf{0.68} & \textbf{0.64} & \textbf{0.35} & \textbf{0.71} & \textbf{0.42} & \textbf{0.41} & \textbf{0.49} & \textbf{0.54} & \textbf{0.77} & \textbf{0.56} \\
        \hline
        Falcon 7B & 0.25 & 0.30 & 0.21 & 0.28 & 0.35 & 0.20 & 0.25 & 0.31 & 0.44 & 0.29 \\
        Falcon 7B Instruct & 0.31 & 0.32 & 0.23 & 0.30 & 0.35 & 0.22 & 0.24 & 0.34 & 0.36 & 0.30 \\
        Falcon 40B & 0.28 & 0.35 & 0.23 & 0.40 & 0.34 & 0.22 & 0.26 & 0.29 & 0.45 & 0.31 \\
        Falcon 40B Instruct & 0.30 & 0.33 & 0.25 & 0.40 & 0.34 & 0.24 & 0.30 & 0.31 & 0.45 & 0.32 \\
        \hline
    \end{tabular}
    \caption{Performance of various models over the AraSTEM dataset. The results are reported per subject}
    \label{table:zero_shot_performance}
\end{table*}

Table \ref{table:zsp-level} presents the accuracy of selected models across different educational levels. The results indicate that the dataset is challenging, even at the primary level. Moreover, as the educational level increases, the models' performance declines, suggesting that higher-level questions are  more demanding.

\subsection{Models' Characteristics}

\begin{figure}[H]
\centering
\captionsetup{justification=centering,margin=0.1cm}
\includegraphics[width=0.5\textwidth]{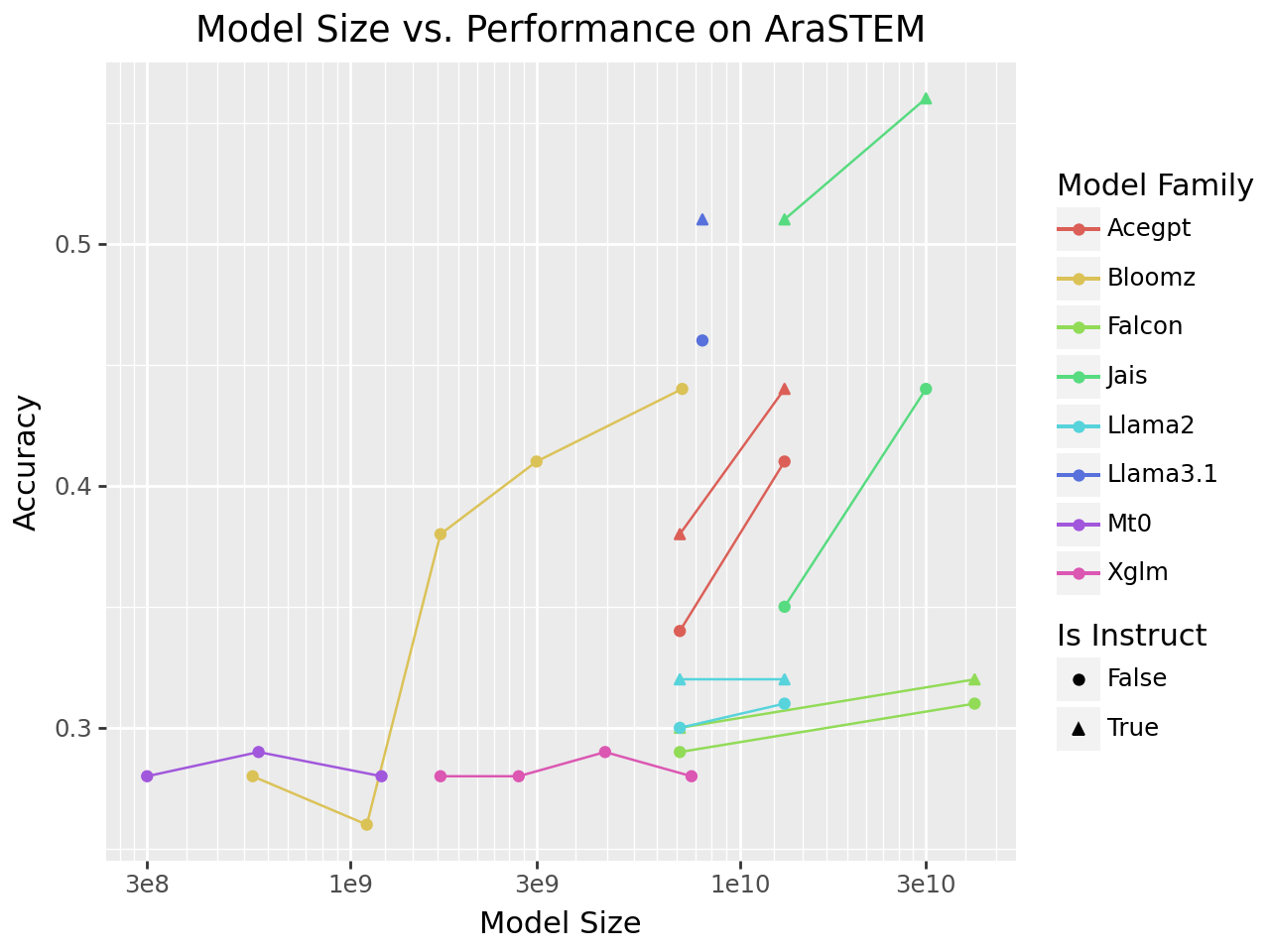}
    \caption{The relationship between model performance on AraSTEM and model size shows that when models are trained on Arabic data, increasing the model size significantly boosts performance}
\label{fig:model_size}
\end{figure}

In this section, we explored how model characteristics affect performance on the AraSTEM dataset. Specifically, we examined two key characteristics: model size and whether the model was fine-tuned for instruction following. Figure \ref{fig:model_size} illustrates the relationship between these characteristics and performance on AraSTEM. The figure shows how changes in model size and the presence of instruction fine-tuning influence performance.

The figure reveals that three models - Jais, Bloomz, and AceGPT - only experience a performance boost when their model size increases. As mentioned in section \ref{sec:zero-shot}, these models have incorporated Arabic into their training datasets \cite{Sengupta2023, Muennighoff2022, huang-etal-2024-acegpt}. In contrast, other models like XGLM, Llama 2, and Falcon do not show a significant improvement in performance as their size increases. It's worth noting that Llama 3.1 achieved a high score despite its relatively smaller size. We hypothesize that this success is likely due to the model being trained on a substantial amount of Arabic tokens, even though this is not explicitly stated in the model's paper \cite{grattafiori2024llama3herdmodels}.

 The trend presented in the figure can be explained by the well-known scaling law for large language models, known as the Chinchilla law \cite{Hoffmann2022}, which suggests that for optimal performance, we should scale the dataset size alongside the model size. In our case, this would mean scaling the inclusion of Arabic in the training data to see better results as the model size grows. 
 
\begin{figure}[H]
\centering
\captionsetup{justification=centering,margin=0.1cm}
\includegraphics[width=0.5\textwidth]{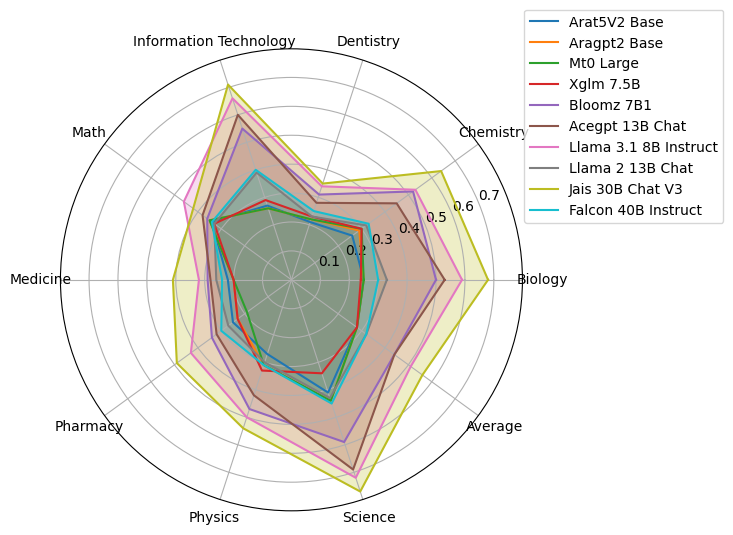}
    \caption{A radar plot to show the performance of selected models per subject on the AraSTEM dataset}
\label{fig:radar}
\end{figure}

Among all the models, Llama 3.1 and Bloomz stands out for achieving high accuracy despite having a smaller size. Bloomz 7B1, for example, performs on par with AceGPT 13B. The key difference is that Bloomz has not been fine-tuned for instructions.Regarding the presence of instruction fine-tuning, Jais shows a significant performance boost between its chat models (which are fine-tuned for instructions) and its raw models. Also Llama 3.1 shows a similar trend but to a lesser extent. While the other models show a modest performance boost.

\subsection{Models' Calibration}

\begin{figure}[H]
\centering
\captionsetup{justification=centering,margin=0.1cm}
\includegraphics[width=0.5\textwidth]{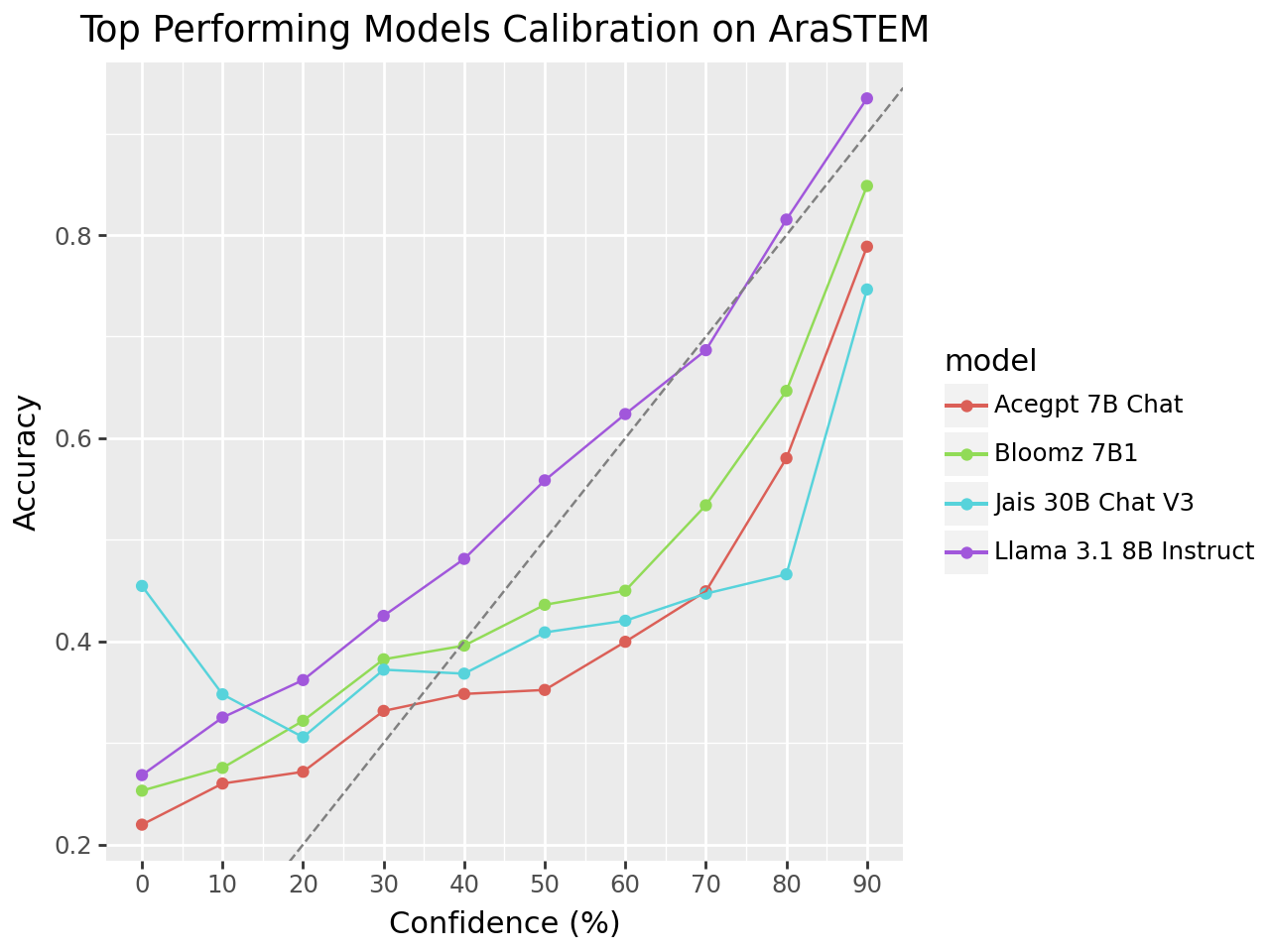}
    \caption{The calibration plot of the top performing models on AraSTEM. The models are over confident with their predictions}
\label{fig:calibration}
\end{figure}

In this section we study the models' confidence against its accuracy. To investigate, we computed calibration plots for the four top-performing models: Jais, Llama 3.1,  Bloomz, and AceGPT. Figure \ref{fig:calibration} illustrates these plots. The results show that for predictions with high confidence, the actual accuracy falls below the perfect calibration line for all models except for Llama 3.1. This indicates that the models tend to be overconfident in their predictions, highlighting the need for calibration to better match their confidence levels with their true performance. On the other hand, Llama 3.1 shows a calibrated behavior as it follows the perfect calibration line for predictions with high confidence.

\subsection{Analysis of Hard Failures}

We analyzed the set of questions that none of the tested models could answer correctly. This subset is relatively small, consisting of only 193 questions (0.16\% of the dataset), predominantly in dentistry and medicine, with a few questions in math and science. This indicates that the models' predictions are complementary, as each model excels in answering different questions.

\begin{figure}[H]
\centering
\captionsetup{justification=centering,margin=0.1cm}
\includegraphics[width=0.5\textwidth]{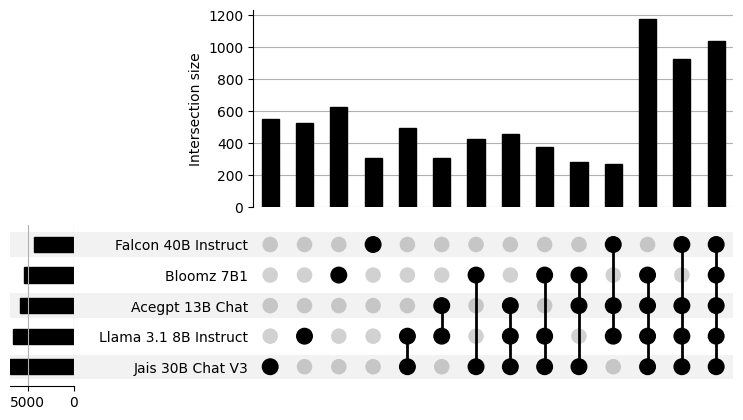}
    \caption{An upset plot to show the intersection of the correct answers figured out by five models on the AraSTEM dataset}
\label{fig:upset}
\end{figure}

To explore this further, we created an upset plot to examine the overlap in correctly answered questions among five models: Jais, Bloomz, AceGPT, Falcon, and Llama 3.1. Figure \ref{fig:upset} displays the intersections with more than 250 shared correct answers.

The plot reveals that while the five models share around 1000 questions that have been answered right, we still have a number of questions that have been answered by solo models. For example, Bloomz 7B1 uniquely solved about 600 questions that no other model answered, while Jais uniquely solved over 550 questions. This demonstrates the complementary nature of the models' predictions, with different models contributing unique strengths.

\subsection{Analysis of Subject-Wise Understanding}

Figure \ref{fig:radar} illustrates the performance of selected models on the AraSTEM dataset, broken down by subject. The results reveal three distinct performance clusters:

\textbf{Top-performing models}: Jais, Llama 3.1, Bloomz, and AceGPT compose this cluster, following a similar performance pattern. These models struggle with the dentistry dataset but excel in information technology, showing a noticeable performance spike. Among them, Jais stands out for its superior performance in the medicine and pharmacy subjects. Llama 3.1 slightly beats Jais in math.

\textbf{Moderate performers}: Falcon and Llama 2 belong to this cluster, also displaying a similar performance trend. Both achieve relatively good results in biology and information technology but generally lag behind the top models.

\textbf{Low performers}: The final cluster comprises models that perform close to random guessing, indicating significant challenges in handling the dataset.

These patterns highlight the strengths and limitations of current models across various STEM subjects.

\section{Conclusion} \label{conclusions}

To address the lack of benchmarks for evaluating LLM knowledge in Arabic, we introduced in this paper, the AraSTEM dataset. It comprises 11,637 multiple-choice questions across STEM subjects such as information technology, math, physics, chemistry, biology, and medicine. AraSTEM has proven to be both diverse and challenging for state-of-the-art open-source Arabic and non-Arabic LLMs. The results highlight the importance of training on substantial amounts of Arabic text for achieving strong performance on Arabic benchmarks, emphasizing the need for greater localization in selecting training datasets. As a future work we aim to investigate applying explainability techniques \cite{elzini} on best performing models. 

\section{Acknowledgments}

We would like to extend our gratitude to the University Research Board and the AI, Data Science, and Computing Hub at the American University of Beirut for funding this research. Our thanks also go to the Maroun Semaan Faculty of Engineering and Architecture (MSFEA) at AUB for providing the essential infrastructure needed to conduct this study.

\bibliographystyle{ieeetr} 
\bibliography{bibliography.bib}
\newpage

\onecolumn
\section{Appendix}
\subsection{Word Cloud} \label{app:word clouds}

In this section, we present the results of generating a word cloud for a number of subjects in the AraSTEM dataset. For each subject, the text from all questions was concatenated and used to create the corresponding word cloud. These visualizations clearly demonstrate that each subject has a distinct focus. For example, recurring words in the physics word cloud include "electricity," "resistance," and "current," while in dentistry, common terms such as "teeth," "tooth," "gum," and "cure" dominate.

\begin{figure*}[htbp]
    \centering
    \subfloat[Math Word Cloud]{\includegraphics[width=0.45\textwidth]{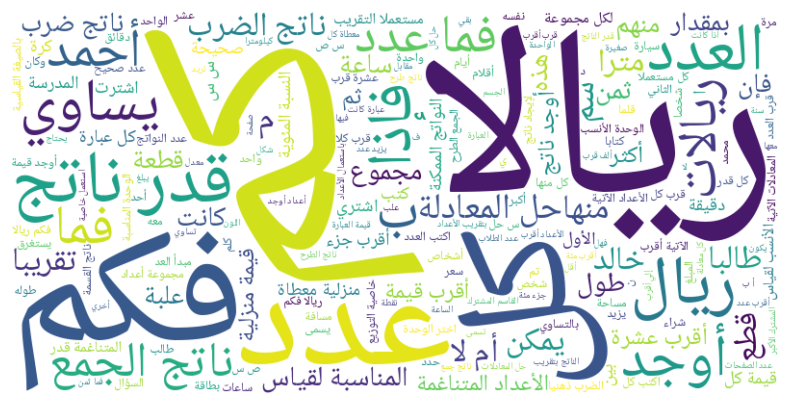}} \hfill
    \subfloat[Physics Word Cloud]{\includegraphics[width=0.45\textwidth]{physics_word_count.png}} \\
    \subfloat[Chemistry Word Cloud]{\includegraphics[width=0.45\textwidth]{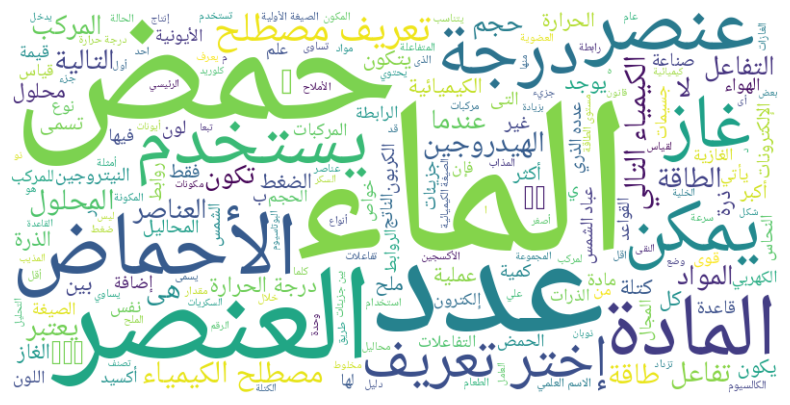}} \hfill
    \subfloat[Dentistry Word Cloud]{\includegraphics[width=0.45\textwidth]{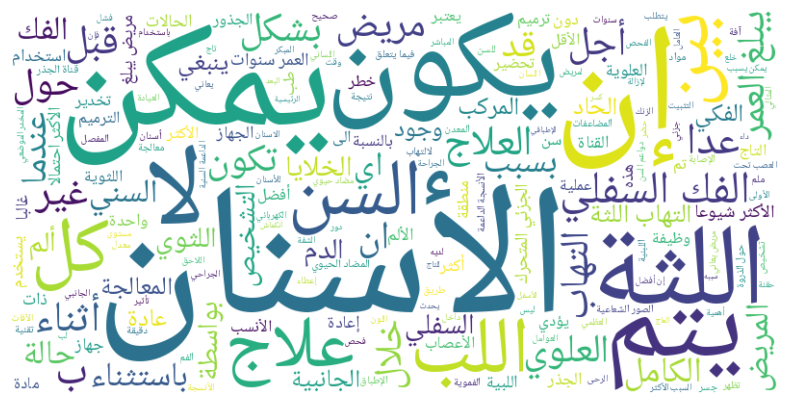}} \\
    \subfloat[Pharmacy Word Cloud]{\includegraphics[width=0.45\textwidth]{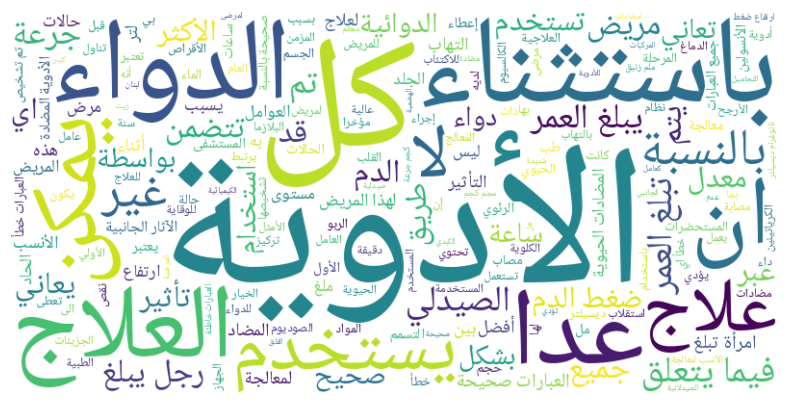}} \hfill
    \subfloat[Medicine Word Cloud]{\includegraphics[width=0.45\textwidth]{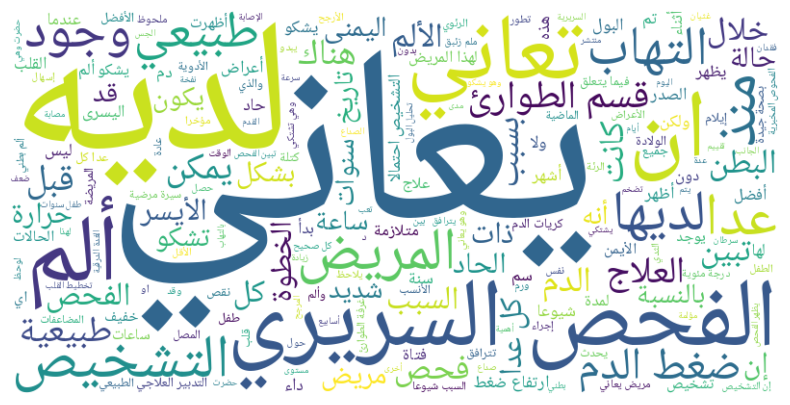}}
    \caption{AraSTEM questions word cloud for each subject}
    \label{fig:grid_images}
\end{figure*}

\end{document}